\documentclass{comjnl}

\usepackage{amssymb}
\usepackage{amsmath}
\usepackage{subcaption}
\usepackage{breqn}
\usepackage[ruled,vlined,algo2e]{algorithm2e}
\usepackage{algorithm,algpseudocode}
\usepackage{booktabs}
\usepackage{url}

\begin{document}

\title{ALGAN: Time Series Anomaly Detection with Adjusted-LSTM GAN}
\author{Md Abul Bashar, Richi Nayak}
\affiliation{School of Computer Science, Faculty of Science, Queensland University of Technology, Brisbane, QLD 4000, Australia\\
ORCID: 0000-0003-1004-4085, 0000-0002-9954-0159} \email{m1.bashar@qut.edu.au, r.nayak@qut.edu.au}

\shortauthors{MA Bashar, R Nayak}
 
\received{00 January 2009}
\revised{00 Month 2009}


\begin{abstract}
Anomaly detection in time series data, to identify points that deviate from normal behaviour, is a common problem in various domains such as manufacturing, medical imaging, and cybersecurity. Recently, Generative Adversarial Networks (GANs) are shown to be effective in detecting anomalies in time series data. The neural network architecture of GANs (i.e. Generator and Discriminator) can significantly improve anomaly detection accuracy. In this paper, we propose a new GAN model, named Adjusted-LSTM GAN (ALGAN), which adjusts the output of an LSTM network for improved anomaly detection in both univariate and multivariate time series data in an unsupervised setting. We evaluate the performance of ALGAN on 46 real-world univariate time series datasets and a large multivariate dataset that spans multiple domains. Our experiments demonstrate that ALGAN outperforms traditional, neural network-based, and other GAN-based methods for anomaly detection in time series data.
\end{abstract}

\maketitle

\section{Introduction}
The proliferation of the Internet of Things (IoT) and the increasing reliance on networked sensors and actuators in various contexts, such as smart buildings, factories, power plants, and data centres, have led to the generation of large quantities of time series data. By continuously monitoring the functioning of these systems, anomalies or behaviour patterns in time steps that deviate from what is considered normal \cite{Chandola2008ComparativeData}, can be detected \cite{Chen2022EffectivelyModel}. Anomaly detection allows for proactive intervention to address underlying issues before they result in potentially catastrophic outcomes \cite{Nagao2013Time-SeriesJapan,Wambura2022RobustSeries}, particularly in some environments that are mission-critical. It is important to note that the ability to accurately identify anomalies is crucial in several domains. It can prevent disruptions in service and ensure the safe and efficient operation of these systems \cite{Bashar2020TAnoGAN:Networks}.

Due to the lack of label information, anomaly detection in time series data is often approached as an unsupervised machine learning problem \cite{Li2019,Schlegl2017UnsupervisedDiscovery,DiMattia2019ADetection}. Most traditional unsupervised anomaly detection methods utilise linear projection and transformation techniques \cite{Li2014ATransform,wold1985encyclopedia}. However, these methods are not well-suited for handling non-linear interactions inherent in time series data. Another class of methods compares the system state value at the current time step with a predicted normal range to identify anomalies \cite{DiMattia2019ADetection}. However, these methods are often ineffective in practice because many systems are highly dynamic and it can be difficult to define a normal range of measurements.

Generative Adversarial Networks (GANs) are gaining popularity as a framework for learning generative models through adversarial training \cite{GoodfellowIanandPouget-AbadieJeanandMirzaMehdiandXuBingandWarde-FarleyDavidandOzairSherjilandCourvilleAaronandBengio2014}. GANs have been successful in generating realistic images \cite{GoodfellowIanandPouget-AbadieJeanandMirzaMehdiandXuBingandWarde-FarleyDavidandOzairSherjilandCourvilleAaronandBengio2014,Denton2015DeepNetworks,Donahue2016AdversarialLearning} and synthetic data \cite{Esteban2017Real-valuedGANs,Mogren2016C-RNN-GAN:Training}, and have been used to model the complex, high-dimensional distribution of normal time series data in the real world \cite{Esteban2017Real-valuedGANs}. An emerging trend in anomaly detection is to combine GANs with latent space mapping to identify anomalies in the data \cite{Bashar2020TAnoGAN:Networks,Schlegl2017UnsupervisedDiscovery,Li2019,Esteban2017Real-valuedGANs}. However, it is yet to be seen how the internal network of GANs can impact the accuracy of detection in time series data.

In this paper, we propose a novel model called Adjusted-LSTM GAN (ALGAN) for detecting anomalies in time series data. It uses an Adjusted-LSTM (ALstm) where the output of the LSTM (Long Short-Term Memory) network is adjusted using attention on top of inputs and hidden states to  enhance the temporal dependencies among the
inputs and the hidden states. This reduces the information loss during long-sequence information propagation through LSTM and, hence, improves anomaly detection in both univariate and multivariate time series data in an unsupervised setting. To detect anomalies in time series data using GANs, it is necessary to model the normal behaviour of the data using adversarial training, and then use an anomaly score to identify points that deviate significantly from this normal behaviour \cite{Schlegl2017UnsupervisedDiscovery,Zenati2018EfficientDetection,Li2019}. To learn this anomaly score, we map the given time series data to a latent space and then reconstruct the data from this latent space. The anomaly score is calculated as the loss between the given data (i.e. real) and the reconstructed (i.e. fake) data.

We evaluated the performance of ALGAN on 47 real-world datasets originating from various domains, 46 of which are univariate time series datasets from the Numenta Anomaly Benchmark (NAB) collection \cite{Lavin2017,Lavin2015EvaluatingBenchmark}, each containing a small number of data points ranging from one thousand to 22 thousand. The remaining dataset is a large multivariate dataset from the Secure Water Treatment (SWaT) system \cite{Mathur2016SWaT:Security}, containing over 946K data points. We compared the performance of ALGAN to that of traditional models, neural network models and other GAN-based methods. Empirical analysis shows that ALGAN significantly outperformed these models. These results demonstrate the effectiveness of ALGAN and the use of ALstm in ALGAN for anomaly detection, as compared to existing GANs.

The main contributions of this paper are:
\begin{enumerate}
    \item The introduction of ALGAN, a novel method based on Adjusted-LSTM, for detecting anomalies in both univariate and multivariate time series datasets.
    
    \item The development of a novel model, Adjusted-LSTM, for adjusting the output of LSTM networks for reducing information loss and enhancing the temporal dependencies among the inputs and the hidden states.
    
    \item An extensive evaluation of ALGAN on a diverse range of 47 univariate and multivariate time series datasets from multiple domains. Results demonstrate that ALGAN, which utilises Adjusted-LSTM, outperforms traditional, neural network-based, existing LSTM-based GANs and other GAN-based approaches.
\end{enumerate}

To the best of our knowledge, this is the first GAN-based work that adjusts the output of LSTM networks for detecting anomalies in long sequences. Examining the feasibility of using ALstm and GAN for anomaly detection in time series data will enable the application of ALstm and GAN feasible in many domains. We show the generalisation ability of our ALstm and GAN framework on various datasets with different characteristics.

The rest of the paper is structured as follows: Section \ref{sec:related_work} summarises related work in the field, Section \ref{sec:TAnoGAN} presents the proposed ALGAN method, Section \ref{sec:evaluation} provides an empirical evaluation of the proposed model, and Section \ref{sec:conclusions} includes the paper conclusion.

\section{Related Work}
\label{sec:related_work}
Principal Component Analysis (PCA) \cite{Li2014ATransform} and Partial Least Squares (PLS) \cite{wold1985encyclopedia} are two widely used linear model-based methods for unsupervised anomaly detection that aim to reduce the dimensionality of high-dimensional data and extract latent features. However, these methods have the limitation of requiring Gaussian distribution and high correlation in the data \cite{Dai2013FromDiagnosis}, which may not be suitable for complex real-world scenarios. K-Nearest Neighbor (KNN) is a widely used method for unsupervised anomaly detection that relies on distance measures \cite{Angiulli2002FastSpaces}. 
However, a limitation of distance-based methods is that they require prior knowledge of the duration and the number of anomalies in the data. Angle-Based Outlier Detection (ABOD) \cite{Kriegel2008Angle-basedData} and Feature Bagging (FB) \cite{Lazarevic2005FeatureDetection} are probabilistic models based on density estimation that show better performance than distance-based methods. However, a drawback of these methods is that they are unsuitable for time series data as they ignore the temporal correlations \cite{Li2019}.

Gaussian Mixture Model (GMM) \cite{bishop2006linear,Reynolds2009}, Isolation Forest (IsoF) \cite{Liu2008IsolationForest}, and One Class Support Vector Machine (OcSVM) \cite{Scholkopf} are widely used methods for anomaly detection in univariate time series data. In this paper, we present experimental results that demonstrate the effectiveness of GMM over IsoF and OcSVM, and the superiority of neural network-based models over GMM. A brief overview of these methods is provided in Section \ref{sec:baselines}.

Deep learning-based unsupervised anomaly detection methods such as Auto-Encoder \cite{Zhou2017AnomalyAutoencoders}, Encoder-Decoder \cite{Habler2018UsingMessages} and LSTM \cite{Malhotra2015,hochreiter1997long} have attracted attention in recent years due to their high performance. A recent development in this field is using the GAN framework to build anomaly detection models but mostly for the image domain only. GANs can perform distribution fitting by learning the underlying distribution of the data. The ability of GANs to generate data can help alleviate the problem of insufficient abnormal samples. By training with a dataset that contains only normal samples and learning the feature representations in latent space, the abnormal samples, which are poorly reconstructed, can be detected. However, the time-series domain brings additional challenges due to the requirement to include time dependencies.

For example, A GAN-based method enables the learning of generative models that can produce realistic images \cite{GoodfellowIanandPouget-AbadieJeanandMirzaMehdiandXuBingandWarde-FarleyDavidandOzairSherjilandCourvilleAaronandBengio2014,Denton2015DeepNetworks,Donahue2016AdversarialLearning}. 
AnoGan, a GAN-based unsupervised learning framework, performs anomaly detection in medical imaging data \cite{Schlegl2017UnsupervisedDiscovery}. AnoGan leverages an adversarial network to capture the normal anatomical variability in the data and employs an anomaly scoring scheme that maps images to a latent space and reconstructs them. The reconstruction error is used as an anomaly score, which reflects the deviation of an image from the learned data distribution and can identify anomalous images. Since AnoGan utilises CNNs in its generator and discriminator to effectively handle image data, it lacks any mechanism to account for temporal dependencies in the generator and is not found suitable for time series data. In contrast, \cite{Esteban2017Real-valuedGANs} developed RGAN, a GAN-based model, that incorporates LSTM units in its generator and discriminator to generate realistic time series data in the medical domain. However, it does not address the problem of anomaly detection. \cite{Geiger2020TadGAN:Networks} presents TadGAN for unsupervised anomaly detection in time series data based on GANs. Similar to RGAN, it employs LSTM networks as the base models for both the Generators and the Critics (or discriminator). Another similar model is Heterogeneous Temporal Anomaly-reconstruction GAN (HTA-GAN) proposed in \cite{Chen2022EffectivelyModel} where Bidirectional LSTM (BiLSTM) is used. 

Recently, TGAN-AD, a transformer-based GAN \cite{Xu2022TGAN-AD:Data} is proposed for anomaly detection. TGAN-AD uses a similar setting as that of RGAN. However, unlike RGAN, it uses transformers in generator and discriminator. Another transformer-based model is proposed by \cite{Xu2021AnomalyDiscrepancy} called Anomaly Transformer. However, Anomaly Transformer uses self-attention weights of each time point as a measure of its association with the whole series, and then computes the association discrepancy between the prior and serial associations. The association discrepancy is based on the observation that anomalies tend to have weak associations with the whole series, and thus concentrate on their adjacent time points. 

However, the experimental results in \cite{Zeng2022AreForecasting} show that the self-attention mechanism is permutation-invariant and anti-ordering, which means that it does not preserve the order and continuity of time series data effectively. \cite{Zeng2022AreForecasting} 
argues that transformers are not suitable for modelling the temporal relations and dynamics of time series data. 
There are several advantages to using LSTM for time series data over transformer; (a) LSTM can handle variable-length inputs and outputs more naturally without requiring padding or truncating the sequences to a fixed length, (b) LSTM can explicitly preserve the sequential order of inputs and outputs without relying on positional embeddings, (c) LSTM can be more robust to noise and outliers in the data, and
(d) LSTM can be easier to interpret and debug. This allows the ALGAN to identify subtle and complex patterns in the data that may be difficult to capture using other networks. 

The GAN models in \cite{Geiger2020TadGAN:Networks,Li2019,Esteban2017Real-valuedGANs} were proposed for anomaly detection in time series datasets. They uses simple LSTM units in generator and discriminator network. However, simple LSTM units are prone to the problem of vanishing gradients, which means that they may have difficulty in learning long-term dependencies in the data.

Different from these works, our paper introduces a novel GAN-based framework for anomaly detection in time series data called ALGAN which employs adjusted LSTM units as the generator and discriminator models to handle the long-term temporal dependencies in the data. 
 
We benchmark the performance of our ALstm-based GAN model with neural network-based models such as Auto-Encoder \cite{Zhou2017AnomalyAutoencoders}, Encoder-Decoder \cite{Habler2018UsingMessages} and LSTM \cite{Malhotra2015,hochreiter1997long} to illustrate the benefit of ALstm and adversarial training over LSTM and traditional training. Moreover, we show that this architecture can detect anomalies in both small and large datasets, as well as, for both univariate and multivariate data.  
Neural network-based models typically require a large amount of data to adjust their parameters \cite{bashar2018cnn} \cite{Li2019}. 

\section{The Proposed Adjusted LSTM Model}
LSTM networks, a type of recurrent neural network, are well-suited for capturing the local temporal characteristics of a sequence \cite{Lim2021TemporalForecasting}. They are designed to retain information from previous time steps and use it for processing the current time step. This allows them to effectively model the dependencies between elements in the sequence and capture the local temporal patterns present in the data. While LSTMs can retain information over longer periods than other types of recurrent networks, they may still struggle to capture very long-term dependencies \cite{Calin2020DeepArchitectures} due to the limitations of the gating mechanism and the finite size of the cell state. 
To overcome this limitation, we propose a method for adjusting the output of the LSTM layer using a self-attention based weight updating mechanism. This proposed architecture is called Adjusted-LSTM (ALstm). In the following sections, we will describe the internal mechanism of LSTMs, attention mechanisms, and the proposed ALstm in more detail.

\subsection{Internal Mechanisms of LSTM}
Figure \ref{LSTM} shows an LSTM cell. In the first stage of the LSTM process, the forget gate employs a sigmoid function to determine  which information from the previous cell state to retain or discard, using both the preceding cell state ($h_{t-1}$) and the current input ($x_t$). The sigmoid function maps the input to a value between 0 and 1, where a value of 0 indicates the removal of the data and a value of 1 indicates its retained. 
\[f_t = \sigma(W_f \cdot [h_{t-1}, x_t] + b_f)\]

In the second stage of the LSTM process, the input gate updates the cell state with new information. This step consists of two parts:
\begin{enumerate}
    \item A sigmoid gate layer is employed to determine which aspects of the cell state should be updated. The sigmoid function maps the input to a value between 0 and 1, where 0 indicates that a particular aspect of the cell state should not be updated, and 1 indicates that it should be updated.
    \[i_t = \sigma(W_i \cdot [h_{t-1}, x_t] + b_i)\]
    \item A tanh gate layer is employed to create candidate values that can be added to the cell state. The tanh function maps the input to a range of -1 to 1. The resulting candidate values are utilised in conjunction with the output of the sigmoid gate layer to update the cell state.
    \[\tilde{C}_t = \tanh(W_C \cdot [h_{t-1}, x_t] + b_C)\]
\end{enumerate}
By combining these two stages, the LSTM network can selectively update the cell state with new information while retaining important information from the preceding cell state. Here, $b_f$, $b_i$ and $b_C$ are bias vectors. 

In the third stage of the LSTM process, the cell state is updated with the information obtained from the previous two stages. Specifically, the forget gate's output (from stage 1) is utilised to remove certain information from the cell state, whereas the input gate's output (from stage 2) is utilised to add new information. These updates are performed using point-wise multiplication and addition, respectively.

The forget gate's output ($f_t$) is multiplied element-wise with the previous cell state ($C_{t-1}$) to determine which elements of the cell state should be retained or discarded. The tanh gate's output  (i.e. candidate values) is multiplied element-wise with the sigmoid gate's output ($i_t$) to determine which elements of the input and hidden states to use for updating the cell state ($C_t$) and to what extent. 
\[C_t = f_t \odot C_{t-1} + i_t \odot \tilde{C}_t\]
Overall, the third stage allows the LSTM network to update the cell state effectively by selectively retaining and modifying the information it contains.

During the final stage of the LSTM process, the cell state is transformed into the hidden state output ($h_t$). This is accomplished via (1) a sigmoid gate that determines which elements of the cell state should be integrated into the output and (2) a tanh function that scales the output to the range (-1, 1).

\[o_t = \sigma(W_o \cdot [h_{t-1}, x_t]+b_o)\]
\[h_t = o_t \odot tanh(C_t)\]

The sigmoid gate receives input from both the preceding hidden state ($h_{t-1}$) and the current input ($x_t$) and generates an output ($o_t$) that signifies which elements of the cell state should be included in the hidden state output. This output is then multiplied element-wise with the output of the tanh function applied to the cell state ($\tanh(C_t)$), resulting in the final hidden state output ($h_t$).

Overall, this final stage allows the LSTM network to produce an updated version of the cell state as its output, allowing it to maintain significant information from the past while also integrating new information as it becomes available.

\subsection{Internal Mechanisms of Self-Attention}
The self-attention mechanism allows a model to focus on important components of the input while processing it \cite{Vaswani2017AttentionNeed}. It works by representing the input as a set of queries, keys, and values. The queries and keys are then used to compute attention scores, which are subsequently employed to weigh the values. Finally, the weighted values are summed to produce the output of the self-attention mechanism. This entire process can be done efficiently using matrix multiplication, as follows. 

\[\text{attention}(Q,K,V) = \text{softmax}(\frac{QK^T}{\sqrt{d_k}})V\]
Where $Q$, $K$ and $V$ are the matrix representation of queries, keys and values respectively. Each of $q \in Q$, $k \in K$ and $v \in V$ is a vector representation of a query, key and value respectively, and $d_k$ is the number of keys. The process is graphically shown in Figure \ref{Attention}.

\subsection{Mechanisms of the proposed Adjusted LSTM} 
 
LSTM networks are effective in modelling long-term dependencies in sequential data, but they face challenges when processing long input sequences. This is because the LSTM compresses all the previous information into a fixed-length vector for each time step, known as the hidden state, during the repeated passes through the LSTM cell. This can lead to information loss or vanishing gradients as the time steps increase and difficulty in remembering long-term dependencies. 

\begin{figure*}[htb!]
    \centering
    \subfloat[Well-established LSTM Network]{{\includegraphics[width=6cm]{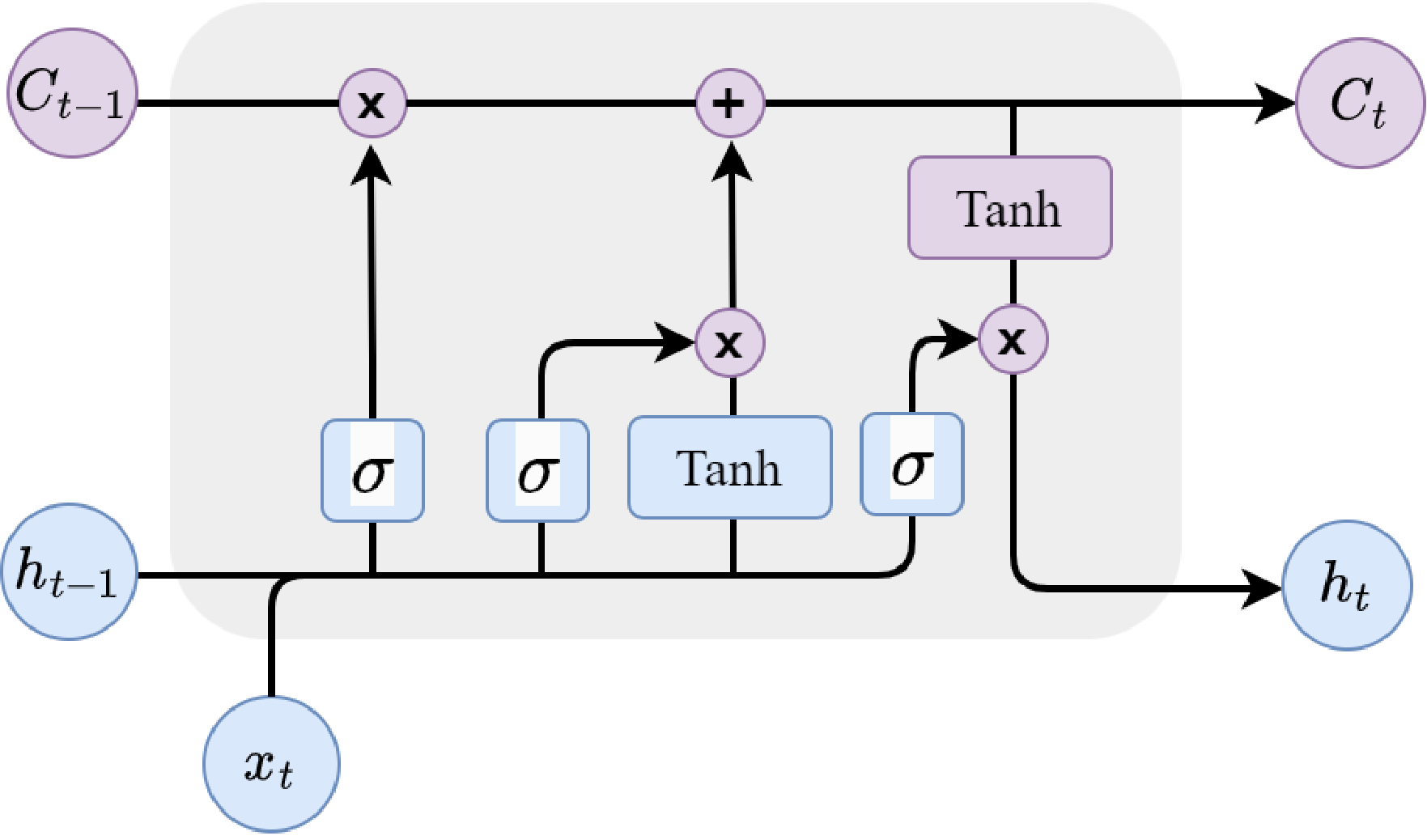}}\label{LSTM}}
    \quad
    \subfloat[Attention]{{\includegraphics[width=3cm]{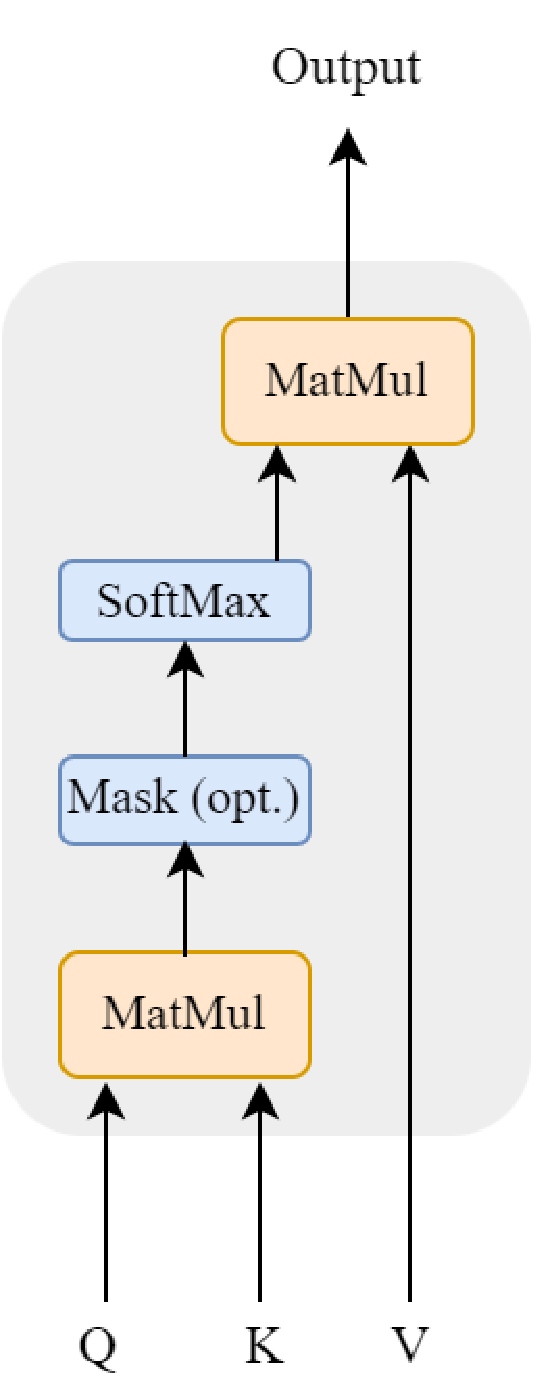}}\label{Attention}}
    \quad
    \subfloat[The proposed Adjusted-LSTM Network]{{\includegraphics[width=4cm]{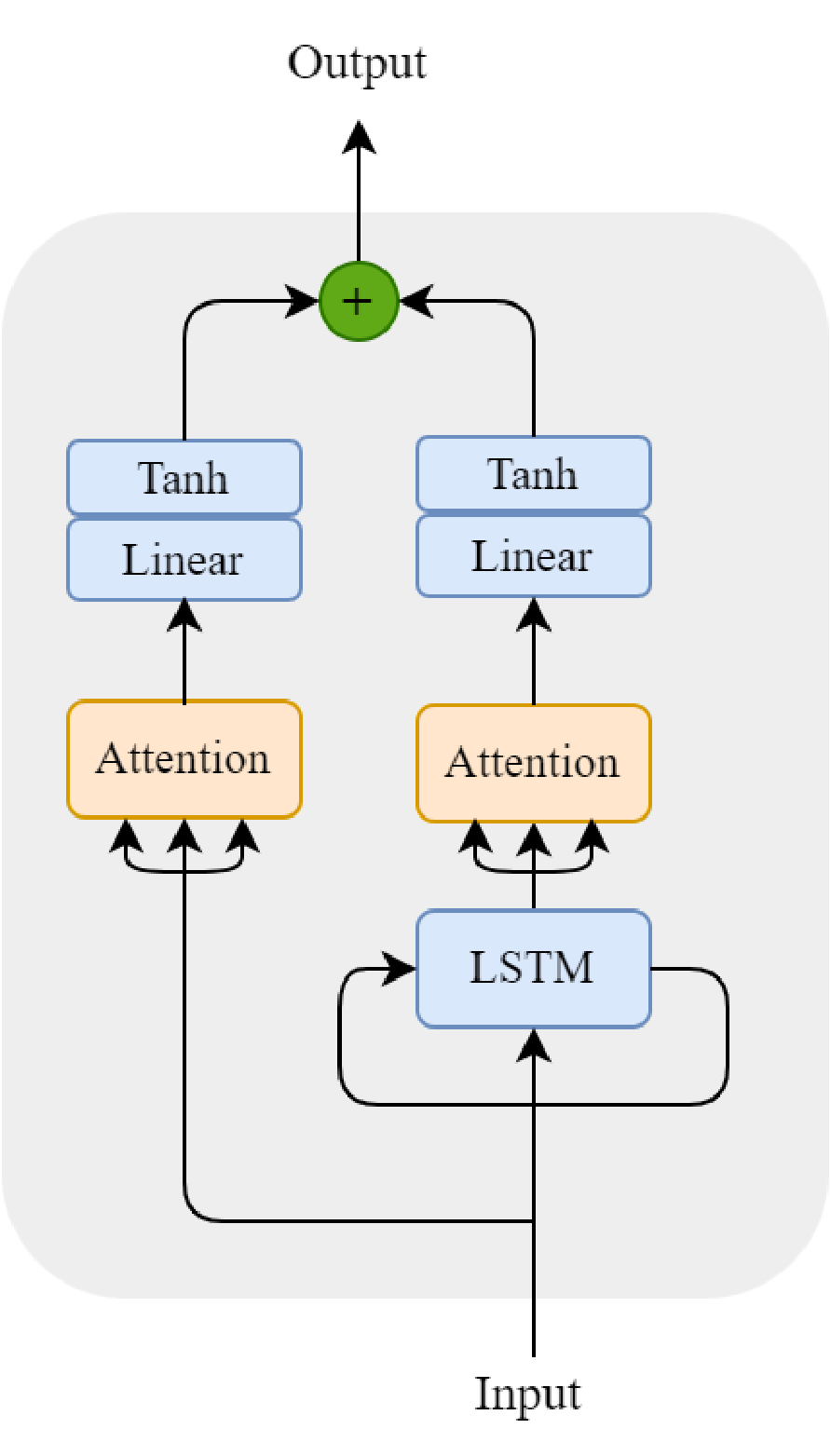}}\label{aLSTM}}
    \caption{Weight Adjusted LSTM suited to handle Time Series data}
    \label{fig:fusion}
\end{figure*}

To address this, we applied the attention mechanism to identify and update the hidden states of time steps in LSTM output that needs adjustment. Figure \ref{aLSTM} shows our proposed Adjusted LSTM (ALstm) network. ALstm consists of two attention layers and two fully connected linear layers with a tanh activation function applied to each layer. 

The first attention layer, in conjunction with the linear layer and tanh activation, is applied on top of the LSTM network to identify the hidden states (of time steps) in the LSTM output that needs adjustment for improving model accuracy. More specifically, the attention mechanism receives input from the hidden states generated by the LSTM at each time step ($h_t$ = $h_1$, $h_2$, $\dots$, $h_N$), which serve as the queries, keys, and values for the attention layer. The attention layer computes attention scores between the queries and keys and uses these scores to weigh the values. The weighted values are then summed to produce the final output of the attention layer ($A^h$ = $a^h_1$, $a^h_2$, $\dots$, $a^h_N$). Optionally, a linear layer with a tanh activation can be added on top of the attention layer to transform the embedding dimension and normalise the values carried from $h_{t}$.

The second attention layer focuses on finding the most pertinent parts of the input sequence that can accurately adjust the hidden states in LSTM output. This involves utilising an attention mechanism to selectively weight the input at each time point ($x_1$, $x_2$, $\dots$, $x_N$) to prioritise important information for updating the attended LSTM output $A^h$. Consequently, the attended output of the input sequence becomes ($A^x = a^x_1$, $a^x_2$, $\dots$, $a^x_N$).

The fully connected linear layer, along with the tanh activation, prepares the chosen input parts to modify the LSTM output. After obtaining the attended output of the input sequence ($A^x$), we feed it through a fully connected linear layer with a tanh activation. The linear layer projects the attended input sequence ($A^x$) to an appropriate dimension needed for updating the attended output of the LSTM ($A^h$). The tanh activation helps to compress the output values within the range [-1, +1], which is necessary for the addition operation to selectively add or remove information. We then perform pointwise addition ($A^x + A^h$) between the attended output of the input sequence ($A^x$) and the attended output of the LSTM ($A^h$) to get the adjusted output of the LSTM unit.  

\section{Generative Adversarial Representation Learning to Identify Time Series Anomalies}
\label{sec:TAnoGAN}

\begin{figure*}[htb!]
    \centering
    \subfloat[Representing Normal Time Series with Generator $G$]{{\includegraphics[width=7.5cm]{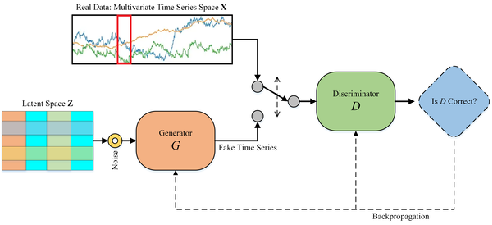}}\label{TAnoGAN_StepOne}}
    \quad
    \subfloat[Mapping Real-Data to the Latent Space and computing Anomaly Scores]{{\includegraphics[width=6cm]{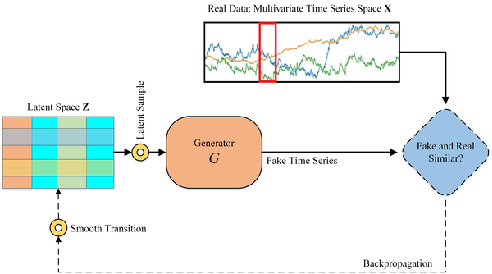}}\label{TAnoGAN_StepTwo}}
    \caption{ALGAN: Time Series Anomaly Detection with Adjusted-LSTM GAN}
    \label{fig:TAnoGAN}
\end{figure*}

The principal objective of unsupervised anomaly detection in time series data is to discern whether the data observations conform to the normal data distribution over time. In cases where the observations do not conform, such non-conforming observations are classified as anomalies \cite{Chalapathy2019DeepSurvey,Kwon2019ADetection}. The proposed ALGAN method involves two sub-processes based on Generative Adversarial Networks (GAN), as illustrated in Figures \ref{TAnoGAN_StepOne} and \ref{TAnoGAN_StepTwo}. In the first sub-process (Figure \ref{TAnoGAN_StepOne}), a generator model is learned, which represents normal time series variability. Concurrently, the generator produce realistic (fake) time series sequences from a latent space. In the second sub-process (Figure \ref{TAnoGAN_StepTwo}), genuine time series sequences are mapped to a latent space and the sequences are reconstructed from the latent space. The reconstruction loss is then utilised to compute anomaly scores and identify anomalies. Algorithm 1 presents a pseudocode for the proposed ALGAN method. The subsequent sections provide details of these sub-processes.

\begin{figure*}[htb!]
    \centering
    \subfloat[Generator $G$ employed in ALGAN]{{\includegraphics[width=6cm]{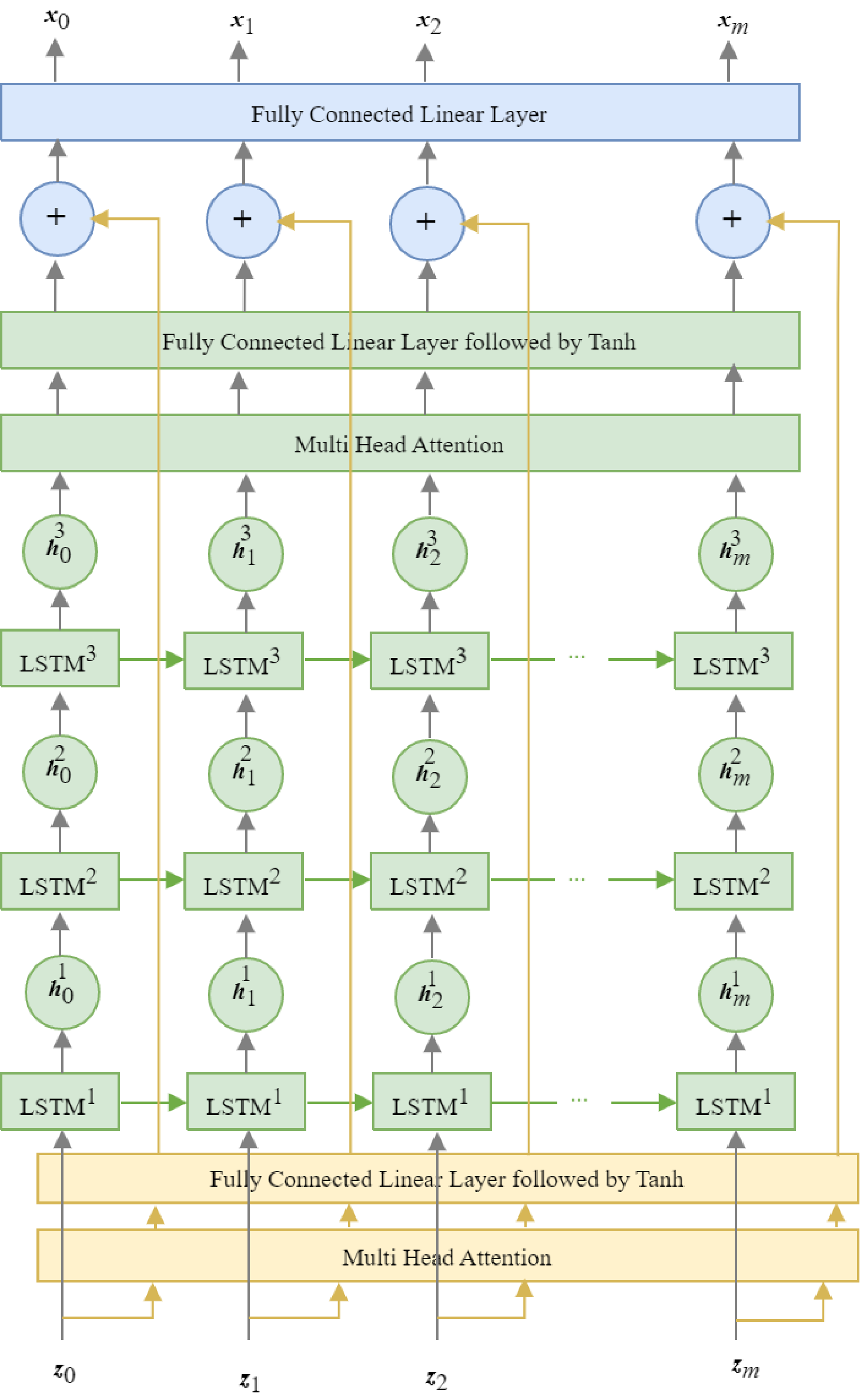}}\label{Generator}}
    \quad
    \subfloat[Discriminator $D$ employed in ALGAN]{{\includegraphics[width=6cm]{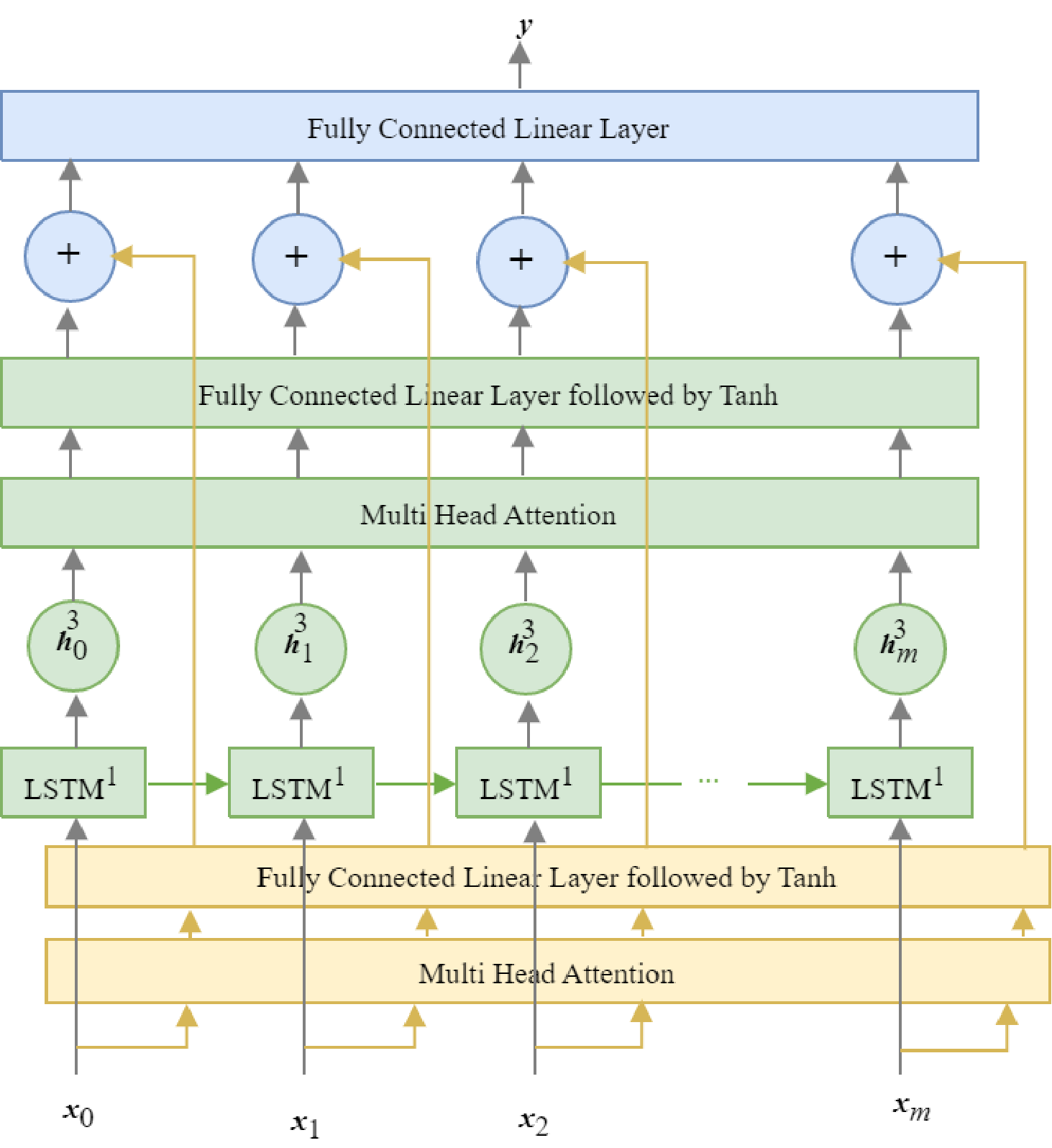}}\label{Discriminator}}
    \caption{Generator and Discriminator Architecture employed in ALGAN}
    \label{Model_Architecture}
\end{figure*}

\subsection{Learning General Data Distribution}
The purpose of this process is to acquire knowledge of the overall data distribution of a dataset via adversarial training. This procedure entails simultaneous training of a Generator $G$, which learns to produce synthetic time series data close to the given data, and a Discriminator $D$, which learns to differentiate between the fabricated data and the actual data. To accommodate time-series data, we introduce the use of our proposed ALstm as a generator model and another ALstm as a discriminator model, as depicted in Figures \ref{Generator} and \ref{Discriminator}, respectively.

To overcome the challenge of insufficient data points, we conducted an in-depth exploration of various architectures for both the generator and discriminator models. We observed that when the size of the dataset is small, a large discriminator model easily overfits the data and a shallow generator model cannot generate realistic data, which is necessary to surpass the discriminator's capabilities. As a remedy, we utilised a shallow discriminator model and a medium-depth generator model. Notably, we found that by employing the ALstm, we can effectively train the generator on both small and large datasets. Driven by empirical analysis, we use the ALstm in the generator model $G$ comprised of three stacked layers with 100 hidden units in each layer and employed a single layer ALstm in the discriminator model $D$ consisting of 100 hidden units in the layer. Figure \ref{Model_Architecture} shows generator and discriminator architecture employed in ALGAN. 

The generator model $G$ takes as input a noise vector $\mathbf{z}$ that is randomly sampled from the latent space $Z$. The original time-series data is partitioned into small sequences using a sliding window $s_w$ before being forwarded to the discriminator model $D$. In parallel, $G$ generates analogous (i.e., fake) small sequences. This process can be conceptualised as a competition between $G$ and $D$, wherein $G$ attempts to generate data that $D$ cannot differentiate from the authentic data. This adversarial relationship between $G$ and $D$ is reminiscent of the GAN framework \cite{GoodfellowIanandPouget-AbadieJeanandMirzaMehdiandXuBingandWarde-FarleyDavidandOzairSherjilandCourvilleAaronandBengio2014}. During this process, $G$ aims to accurately learn the general data distribution, to generate realistic time-series data that $D$ cannot distinguish from the genuine data. The competitiveness between $G$ and $D$ fosters their knowledge and continues until $G$ successfully generates realistic time-series data that is indistinguishable from the authentic data.

In GANs, the function $G(z, \theta_1)$ is responsible for modelling the Generator, which maps input noise vectors $\mathbf{z} \in \mathbf{Z}$ to the desired data space $\mathbf{x} \in \mathbf{X}$, i.e., small time series sequences. On the other hand, the Discriminator is modelled by the function $D(x, \theta_2)$, which outputs the probability that the data is real, where $\theta_1$ and $\theta_2$ are parameters of the models. To train these models, a loss function is employed that maximises the function $D(\mathbf{x})$ and minimises $D(G(\mathbf{z}))$. Through repeated iterations of training, the Generator and Discriminator will eventually converge to a point at which they cannot improve any further, resulting in the Generator producing realistic time series data that the Discriminator is unable to differentiate from real data.

During the training period, both the Generator $G$ and the Discriminator $D$ optimise their respective competitive loss functions. Consequently, they can be viewed as two agents engaged in a minimax game with a value function $V(G,D)$. Specifically, $G$ aims to maximise the probability of $G(\mathbf{z})$ being identified as real by $D$, while $D$ endeavours to minimise this same probability. The value function $V(G,D)$ is defined as follows, following \cite{GoodfellowIanandPouget-AbadieJeanandMirzaMehdiandXuBingandWarde-FarleyDavidandOzairSherjilandCourvilleAaronandBengio2014}.
\begin{dmath}
\min_G \max_D V(D,G) = 
E_{\mathbf{x}\sim p_{data}(\mathbf{x})}[log D(\mathbf{x})] + E_{\mathbf{z}\sim p_z(\mathbf{z})}[log(1-D(G(\mathbf{z})))],
\end{dmath} 
where $E_{\mathbf{x}}$ is the expected value of $\mathbf{x}$ and $E_{\mathbf{z}}$ is the expected value of $\mathbf{z}$.

Stochastic gradient descent (SGD) has been widely adopted as a powerful optimisation technique in various domains \cite{Chalapathy2019DeepSurvey,Kwon2019ADetection,GoodfellowIanandPouget-AbadieJeanandMirzaMehdiandXuBingandWarde-FarleyDavidandOzairSherjilandCourvilleAaronandBengio2014}. In light of its success, we employ the SGD algorithm to train the GAN network. Upon the completion of this adversarial training process, the real time series sequences $\mathbf{x} \in \mathbf{X}$ are projected onto the latent space $\mathbf{z} \in \mathbf{Z}$ for subsequent anomaly detection purposes.

\begin{algorithm}
\scriptsize
\caption{Algorithm for ALGAN}
\SetKwInput{KwInput}{Input} 
\SetKwInput{KwOutput}{Output} 

\DontPrintSemicolon
  \KwInput{A list of small time-series sequences $\mathbf{X}$ and hyper parameter $\Lambda$.}
  \KwOutput{A list of anomaly scores $A$.}

  \SetKwFunction{FMain}{Main}
  \SetKwFunction{FAdvTrn}{adversarial_train}
  \SetKwFunction{FMap}{anomaly_score}
 
    \SetKwProg{Fn}{Function}{:}{}
    \Fn{\FAdvTrn{$\mathbf{X}$}}{
        \For{number\_of\_epochs}{
            Sample $m$ noise vectors $\{\mathbf{z}_1, \dots \mathbf{z}_m\}$ from the noise prior $p_g(\mathbf{z})$.\\
            Generate $m$ fake-data vectors $\{G(\mathbf{z}_1), \dots, G(\mathbf{z}_m)\}$ from the $m$ noise vectors.\\
            Sample $m$ real-data vectors $\{ \mathbf{x}_1, \dots, \mathbf{x}_m \}$ from the data generating distribution $p_{data}(\mathbf{x})$.\\
            Train $D$ on fake-data vectors and real-data vectors.\\
            Sample another $m$ noise vectors $\{\mathbf{z}_1, \dots \mathbf{z}_m\}$ from the noise prior $p_g(\mathbf{z})$.\\
            Train $G$ on the second set of noise vectors. 
        }
        \KwRet $G$, $D$
    }
  \;
    \SetKwProg{Fn}{Function}{:}{}
    \Fn{\FMap{$\mathbf{X}$, $G$, $D$}}{
        \For{$i$ in 1 to $m$}{
            Sample a noise vector $\mathbf{z}^i$ from the noise prior $p_g(\mathbf{z}^i)$.\\
            \For{$\lambda$ in 1 to $\Lambda$}{
                Generate a fake-data vector $G(\mathbf{z}^i)$ from the noise vector $\mathbf{z}^i$.\\
                Calculate Loss $\mathcal{L}(G(\mathbf{z}^i))$ for $\mathbf{x}^i$ utilising $G$ and $D$, and update $\mathbf{z}^i$ using gradient descent.
            }
            $A(\mathbf{x}^i)$ = $\mathcal{L}(G(\mathbf{z}^i))$
        }
        \KwRet $A$
    }
    \;
    \SetKwProg{Fn}{Function}{:}{\KwRet}
    \Fn{\FMain{$\mathbf{X}$}}{
        $G$, $D$ = adversarial\_train ($\mathbf{X}$)\\
        A = anomaly\_score ($\mathbf{X}$, $G$, $D$)
    }
\end{algorithm}

\subsection{Mapping Real-Data to the Latent Space}
In the adversarial training framework, the Generator is trained to learn the mapping $G: \mathbf{Z} \rightarrow \mathbf{X}$, which maps a latent space representation $\mathbf{z}$ to a realistic small time series sequence $\mathbf{x}$ that belongs to the distribution $\mathbf{X}$. The aim is to generate samples that are indistinguishable from the real data. Detecting anomalies requires a thorough understanding of the underlying distribution of the data in order to accurately identify and isolate anomalies in the generated samples. Therefore, when detecting anomalies, it is necessary to map a real small time series sequence $\mathbf{x} \in \mathbf{X}$ to the latent space representation $\mathbf{z} \in \mathbf{Z}$ to assess how closely the corresponding latent space generates real small time series sequences. The degree of similarity between $\mathbf{x}$ and $G(\mathbf{z})$ is dependent on the extent to which $\mathbf{x}$ adheres to the data distribution $p_g$, which was utilised during the training of the Generator $G$. However, GAN does not possess an inverse mapping $G^{-1}: \mathbf{X} \rightarrow \mathbf{Z}$ that maps real small time series sequences to the latent space representation. 

Figure \ref{TAnoGAN_StepTwo} illustrates the procedure employed to map real small time series sequences onto the latent space. The task of determining the optimal $\mathbf{z}$ value for a given $\mathbf{x}$ begins with the random sampling of $\mathbf{z}_1 \in \mathbf{Z}$, which is then fed into the pre-trained generator $G$ to generate a fake small sequence $G(\mathbf{z}_1)$. A loss function $\mathcal{L}$ is defined based on the generated small sequence $G(\mathbf{z}_1)$, which provides gradients for updating the parameters of $\mathbf{z}_1$. This results in an updated position $\mathbf{z_2} \in \mathbf{Z}$ that is expected to be a better fit for the given $\mathbf{x}$. To find the $\mathbf{z}$ that produces the most similar small sequence $G(\mathbf{z}\Lambda) \sim \mathbf{x}$, an iterative process is employed via $\lambda = 1, 2, \dots, \Lambda$ backpropagation steps.

In this paper, the task of mapping a given real small time series sequence $\mathbf{x} \in \mathbf{X}$ to the most optimal latent space location $\mathbf{z} \in \mathbf{Z}$ is accomplished by defining a loss function $\mathcal{L}$. 
The loss function $\mathcal{L}$ comprises two distinct components: a residual loss $\mathcal{L}_R$ and a discrimination loss $\mathcal{L}_D$, each serving a specific purpose \cite{Schlegl2017UnsupervisedDiscovery}.

\noindent \textbf{Residual Loss} $\mathcal{L}_R$ is a function that quantifies the point-wise dissimilarity (e.g., values at different timestamps) between a real small sequence $\mathbf{x}$ and its corresponding generated counterpart $G(\mathbf{z}_\lambda)$, as produced by the generator network, and it is defined as follows. 
\begin{equation}
    \mathcal{L}_R(\mathbf{z}_\lambda) = \sum \left | \mathbf{x} - G(\mathbf{z}_\lambda) \right |
\end{equation}

\noindent \textbf{Discrimination Loss} component of the loss function involves the utilisation of a rich intermediate feature representation of the discriminator network. Specifically, the output of a designated intermediate layer $f(\cdot)$ of the discriminator network is utilised to specify the statistical properties of a given input small sequence. This approach enables the discrimination loss to effectively capture the discrepancies between the generated and real small sequences in the latent feature space, thereby enhancing the overall performance of the model. The discriminator loss is defined as follows. 
\begin{equation}
    \mathcal{L}_D(\mathbf{z}_\lambda) = \sum \left | f(\mathbf{x}) - f(G(\mathbf{z}_\lambda)) \right |
\end{equation}
The Loss function $\mathcal{L}$ is defined as a weighted sum of the residual loss and discrimination loss as follows.
\begin{equation}
    \mathcal{L}(\mathbf{z}_\lambda) = (1-\gamma) \cdot \mathcal{L}_R(\mathbf{z}_\lambda) + \gamma \cdot \mathcal{L}_D(\mathbf{z}_\lambda)
\end{equation}

The residual loss $\mathcal{L}_R$ and discrimination loss $\mathcal{L}_D$ are critical components of the loss function, each with a distinct role in the optimisation process. Specifically, while $\mathcal{L}_R$ serves to ensure point-wise similarity between the generated fake small sequence $G(\mathbf{z}_\lambda)$ and the real small sequence $\mathbf{x}$, $\mathcal{L}_D$ enforces the constraint that the generated sequence lies within the manifold $\mathbf{X}$ of the real data. To achieve this, both the generator $G$ and discriminator $D$ networks are utilised to update the parameters of $\mathbf{z}$ via backpropagation, with only the parameters of $\mathbf{z}$ being updated while keeping those of $G$ and $D$ fixed. In this research, we set $\gamma$ based on experimental results of anomaly detection accuracy. 

\begin{figure*}[htb!]
    \centering
    \subfloat[Example 1: Anomaly score $A(\mathbf{x})$ in blue, ground truth $y$ in orange and threshold in red colour]{{\includegraphics[width=6cm]{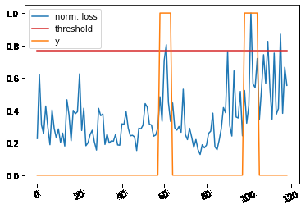}}\label{vis_TAnoGAN_loss1}}
    \quad
    \subfloat[Example 1: System values in green, the ground truth $y$ in orange and detected anomalies in red colour.]{{\includegraphics[width=6cm]{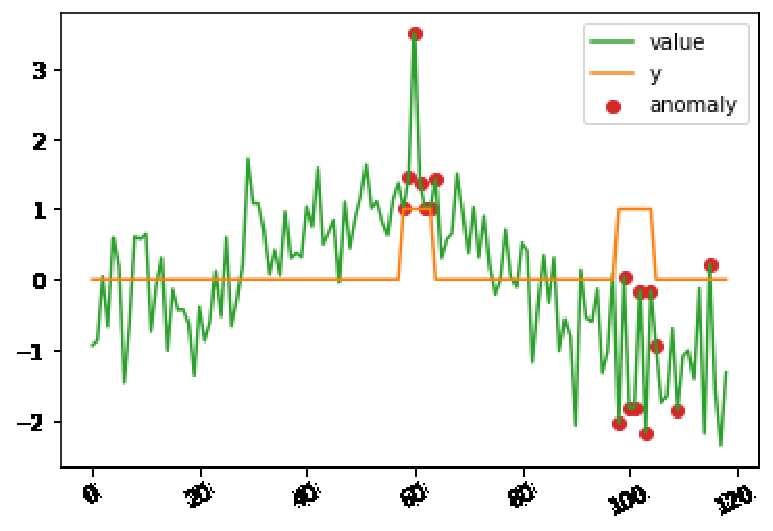}}\label{vis_TAnoGAN_value1}}
    \quad
    \subfloat[Example 2: Anomaly score $A(\mathbf{x})$ in blue, ground truth $y$ in orange and threshold in red colour]{{\includegraphics[width=6cm]{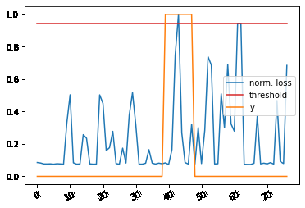}}\label{vis_TAnoGAN_loss2}}
    \quad
    \subfloat[Example 2: System values in green, the ground truth $y$ in orange and detected anomalies in red colour.]{{\includegraphics[width=6cm]{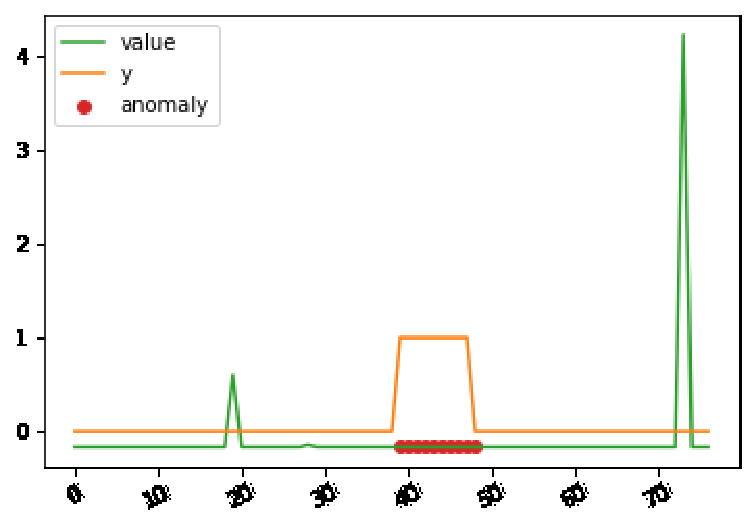}}\label{vis_TAnoGAN_value2}}
    \caption{Visualisation of Anomalies Detection by ALGAN. Note the change in scaling in the Y axis of (a) and (b); as well as (C) and (d)}%
    \label{fig:vis_TAnoGAN}%
\end{figure*}

\subsection{Detection of Anomalies}
When conducting anomaly detection in small sequence time series data, each individual small sequence $\mathbf{x}$ is evaluated and classified as either a normal observation (i.e. it adheres to the general data distribution), or as an anomalous observation (i.e. it deviates from the typical patterns observed in the data). 
 
In the adversarial training regime, the generator $G$ is trained to learn the underlying probability distribution $p_g$ of the latent space $\mathbf{Z}$ as well as $p_{data}$ of the real data space $\mathbf{X}$. Specifically, the generator generates fake small sequences $G(\mathbf{z}_\lambda)$ based on the general data distribution of $\mathbf{X}$. During each iteration, $\lambda$ of the training process, the loss function $\mathcal{L}$ is utilised to assess the dissimilarity between the generated fake small sequence $G(\mathbf{z}_\lambda)$ and the corresponding real small sequence $\mathbf{x}$. The loss function $\mathcal{L}$ can be utilised to derive an anomaly score $A(\mathbf{x})$ that quantifies the degree of fit of a given real small sequence $\mathbf{x}$ to the general data distribution. This anomaly score can be used to identify anomalous or abnormal small sequences in the real data space $\mathbf{X}$ that deviate significantly from the normal data distribution modelled by $p_{data}$. The anomaly score $ A(\mathbf{x})$ is defined as follows. 
\begin{equation}
    A(\mathbf{x}) = (1-\gamma) \cdot \mathcal{R}(\mathbf{x}) + \gamma \cdot \mathcal{D}(\mathbf{x}),
\end{equation}
where the residual score $\mathcal{R}(\mathbf{x})$ and the discrimination score $\mathcal{D}(\mathbf{x})$ are defined by the residual loss $\mathcal{L}_R(\mathbf{z}_\Lambda)$ and the discrimination loss $\mathcal{L}_D(\mathbf{z}_\Lambda)$, respectively, at the final ($\Lambda$th) updating iteration of the mapping procedure to the latent space. The parameter $\gamma$ is a trade-off factor decided empirically. A large anomaly score $A(\mathbf{x})$ indicates an anomalous small sequence whereas a small anomaly score indicates a small sequence fitting to the general data distribution of $\mathbf{X}$ learned by $G$ during adversarial training.

Figure \ref{fig:vis_TAnoGAN} shows two examples of anomalies detected by ALGAN. The blue graphs presented in Figures \ref{vis_TAnoGAN_loss1} and \ref{vis_TAnoGAN_loss2} illustrate the losses pertaining to the re-constructed small sequences, revealing four and one peaks, respectively, over the red threshold lines. The threshold was determined through a combination of visual inspection and model performance evaluation. The orange ground truth lines depicted in these figures indicate that two of the peaks correspond to true positives and the other two are false positives in Example 1, and the peak over the red line is true positive in Example 2. Additionally, Figures \ref{vis_TAnoGAN_value1}(b) and \ref{vis_TAnoGAN_value2} (d) showcase the original time sequences of the dataset through green graphs, while the orange ground truth line illustrates the time spans during which the anomalies originally occurred. The red dots plotted in these figures highlight the points where ALGAN successfully identified the anomalies. Notably, ALGAN can correctly identify anomalies even when the anomaly is not visually apparent. However, it should be noted that there are two instances of false positives in Example 1. Upon closer inspection of the original time series, it is apparent that these two points exhibit a distribution that is distinct from the general trend of the data. 

\section{Empirical Evaluation}
\label{sec:evaluation}
The primary purpose of experiments is to evaluate the performance of the proposed ALGAN model that employs adversarial learning with ALstm model in both generator and discriminator components.

\subsection{Data Collection}
\label{sec:data_collection}

\subsubsection{NAB Dataset Collection}

Numenta Anomaly Benchmark (NAB) data collection \cite{Lavin2017,Lavin2015EvaluatingBenchmark}  is a high-quality repository of time-series data with manually annotated anomalies \cite{Lavin2015EvaluatingBenchmark}. The collection comprises 58 datasets with varying lengths of time series instances ranging from 1000 to 22,000, resulting in a total of 365,551 data points. The collection is grouped into seven categories based on their characteristics and sources, of which five include real-world data and two include artificial data. We use all 46 datasets that belong to the real-world categories for our experiments.
Each record in the NAB time-series datasets consists of a time stamp and a single scalar value. The anomalies in the datasets are marked with a time interval, as they often span over a duration of time rather than occurring at a single point.

We train ALGAN and baseline models on these datasets but without using the labels. The ground truths (i.e., the labels from NAB) are only used to assess the model performance of anomaly detection by comparing the anomalies that are predicted.
We split the original long time-series data into smaller time series using a sliding window for data preparation. The optimal window length is crucial in time-series analysis \cite{Li2019}, and we tried different window sizes to capture the system status at various resolutions. We found that a window size of $s_w = 30 \times 2$ produces reasonably optimal results. To capture the relevant dynamics of the datasets, we use a shift length of 1 during training and testing. A small time series is considered an anomaly if it falls within (or overlaps with) an anomaly period \cite{Lavin2017,Lavin2015EvaluatingBenchmark}. This subdivision allows us to reward early detection and penalize late detection.

\subsubsection{SWaT Dataset}

The Secure Water Treatment (SWaT) system \cite{Mathur2016SWaT:Security} is a testbed for a water treatment plant designed to closely replicate the physical process and control systems of actual field systems in large cities. 
The water treatment process in SWaT
is controlled by a pair of programmable logic controllers (PLCs) that communicate with a variety of sensors (such as water flow indication transmitters, level indicator transmitters, and pH analysers) and actuators (such as pumps and motorized valves). The PLCs collect readings from the sensors to monitor the status of the physical process and send actuation commands to the actuators, which may be used to change or maintain the current state of the actuator based on the PLC's internal logic.

The dataset used in this study was collected from the SWaT testbed over 11 days, with a sampling rate of one record per second. The dataset consists of 946,722 records, each containing 51 attributes that represent readings from 25 sensors and the states of 26 actuators. The dataset has been divided into a training set and a test set. The training set consists of 7 days of normal operation, while the test set includes 4 days of normal operation plus 36 simulated attacks that were generated using the attack model described in \cite{Adepu2016GeneralizedSystems}. 

\subsection{Evaluation Measures}
We used six standard measures to evaluate the performance of anomaly detection: Accuracy (Ac), which is the proportion of correctly classified instances; Precision (Pr), which is the proportion of positive instances that are truly positive; Recall (Re), which is the proportion of positive instances that are correctly identified; F$_1$ Score (F$_1$), which is the harmonic mean of precision and recall; Cohen Kappa (CK), which is a measure of agreement between the predicted and actual labels that account for chance agreement; and Area Under Curve (AUC), which is a measure of how well the model can distinguish between positive and negative instances. A detailed description of these measures can be found in \cite{Bashar2020RegularisingSet}.

\subsection{Baseline Models}
\label{sec:baselines}
We implemented eight state-of-the-art anomaly detection models as baseline models to compare the performance of our proposed ALGAN model. Note all baselines cannot be executed on multivariate datasets, some of them (e.g. traditional models and vanilla LSTM model) are limited to single variate datasets only.

\begin{enumerate}

\item \textbf{Multivariate Anomaly Detection for Time Series Data with GAN (MadGan)} \cite{Li2019}:
In this GAN-based model, the generator consists of a three-layer LSTM network with 100 units in each hidden layer, while the discriminator is a single-layer LSTM network with 100 hidden units.

\item \textbf{Auto Encoder (AutoEn)} \cite{Borghesi2019AnomalySystems,Zhou2017AnomalyAutoencoders}:
This model comprises three layers: an LSTM encoder with 256 units and 20\% dropout, an LSTM decoder with 512 units and 20\% dropout, and a dense output layer with the same number of units as the small time series length. 

\item \textbf{Bidirectional LSTM in GAN (BiLstmGan)} \cite{Zhu2019ElectrocardiogramNetwork,Chen2022EffectivelyModel}:
In this model, the generator is composed of a three-layer bidirectional LSTM network with 50 units in each hidden layer, while the discriminator is a single-layer bidirectional LSTM network with 50 hidden units.

\item \textbf{CNN in GAN (CnnGan)} \cite{Schlegl2017UnsupervisedDiscovery}: 
The generator and the discriminator consist of four stacked CNNs each. Every CNN has two convolutional layers with 20\% dropout and ReLU activation. The kernel size is 256 and the number of units in each convolutional layer matches the small sequence length. The final layer is a dense layer.
\item \textbf{Vanilla Long Short-Term Memory (VanLstm)} \cite{Malhotra2015,hochreiter1997long}:
This model uses LSTM to learn and predict the sequential patterns in the data. A data point is considered anomalous if its actual value deviates significantly from the LSTM prediction based on the previous window of time series data. This model has two LSTM layers with 256 and 512 hidden units respectively, and each layer applies 20\% dropout randomly.
 
\item \textbf{Isolation Forest (IsoF)} \cite{Liu2008IsolationForest}: 
This conventional model randomly chooses a feature and a split point within the range of the chosen feature. A tree structure can represent the recursive partitioning process. The path length from the root to the leaf node corresponds to the number of splits required to isolate a sample. The average path length over a forest of random trees is an indicator of normality. Anomalies tend to have shorter paths due to partitioning. A sample with a short path length is likely to be anomalous.

\item \textbf{Gaussian Mixture Model (GMM)} \cite{bishop2006linear,Reynolds2009}: 
This is the most popular Unsupervised Clustering method that fits $K$ Gaussian curves to the data with $K$ clusters. Each Gaussian has three parameters: mean $\mu$, variance $\sigma$, and mixing probability $\pi$. The mean determines the center, the variance determines the width, and the mixing probability determines the size of the Gaussian. The mixing probabilities sum to one: $\sum_1^K \pi_k = 1$. Maximum likelihood estimates the optimal parameters for each Gaussian based on the data points in each cluster. Gaussian Mixture Models assign a probability $p$ to each data point being generated by one of $K$ Gaussians. A data point is an outlier if $p < \tau$ for all clusters, where $\tau$ is the threshold.

\item \textbf{One Class Support Vector Machine (OcSVM)} \cite{Scholkopf}: 
This unsupervised method finds the minimum hypersphere that can contain most of the data instances. The anomaly score of an instance outside the hypersphere is proportional to its distance from the hypersphere.

\end{enumerate}

\subsection{Accuracy Performance on the NAB Dataset}
Table \ref{tab:addlabel} illustrates the average performance metrics of various models on a collection of 46 datasets in NAB. The results demonstrate that ALGAN surpasses all baseline models on all accuracy measures. It is worth noting that ALGAN outperforms the second-best performing model AutoEncoder  by a clear margin, e.g., 7.12\%, 6.461\%, and 8.224\% for precision, F$_1$ score, and CK metrics, respectively.

\begin{figure*}[htb!]
    \centering

    \subfloat{{\includegraphics[width=4cm]{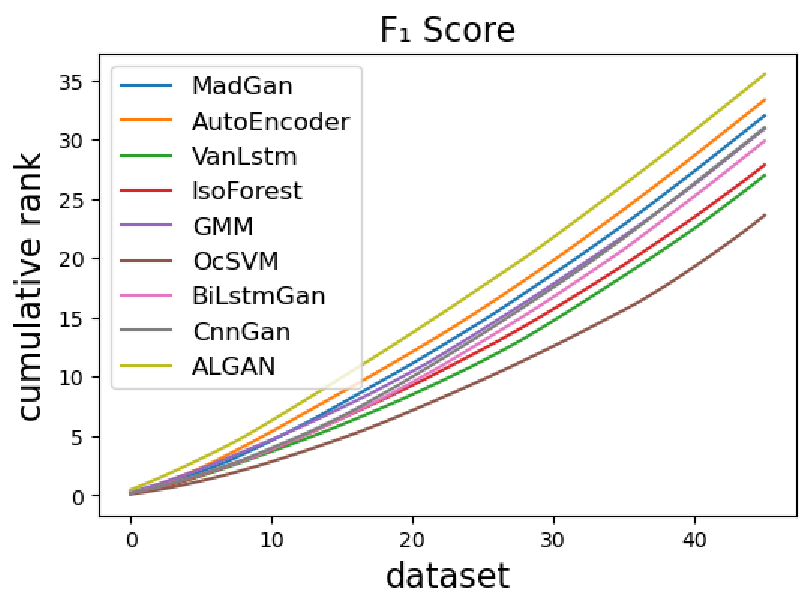}}}
    \quad
    \subfloat{{\includegraphics[width=4cm]{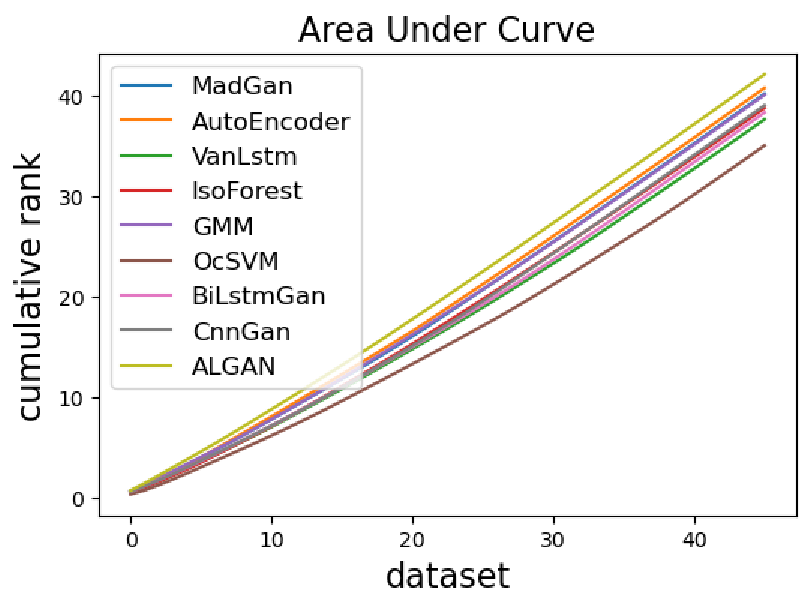}}}
    \quad
    \subfloat{{\includegraphics[width=4cm]{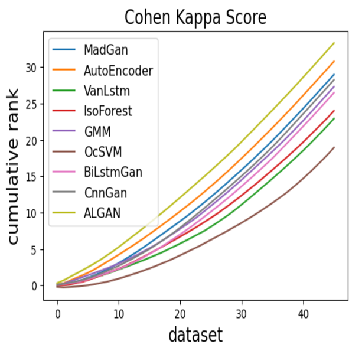}}}
    
    \subfloat{{\includegraphics[width=4cm]{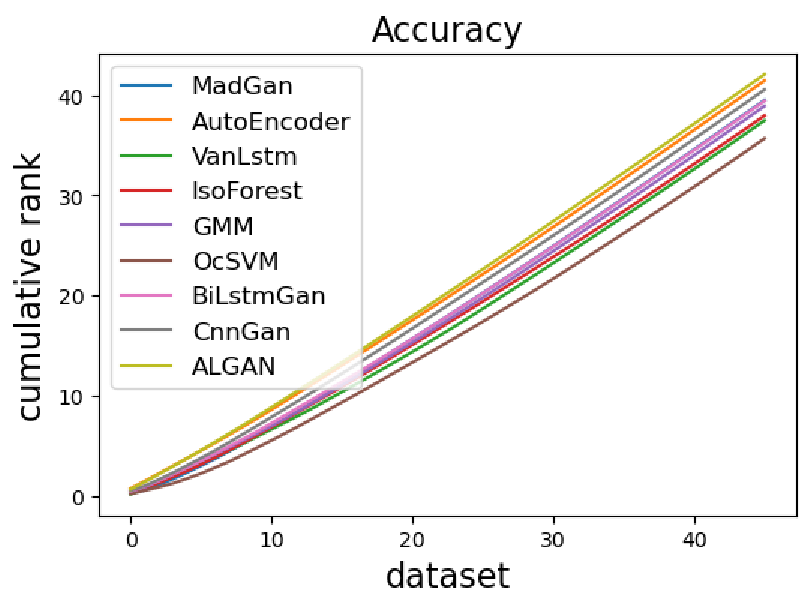}}}
    \quad
    \subfloat{{\includegraphics[width=4cm]{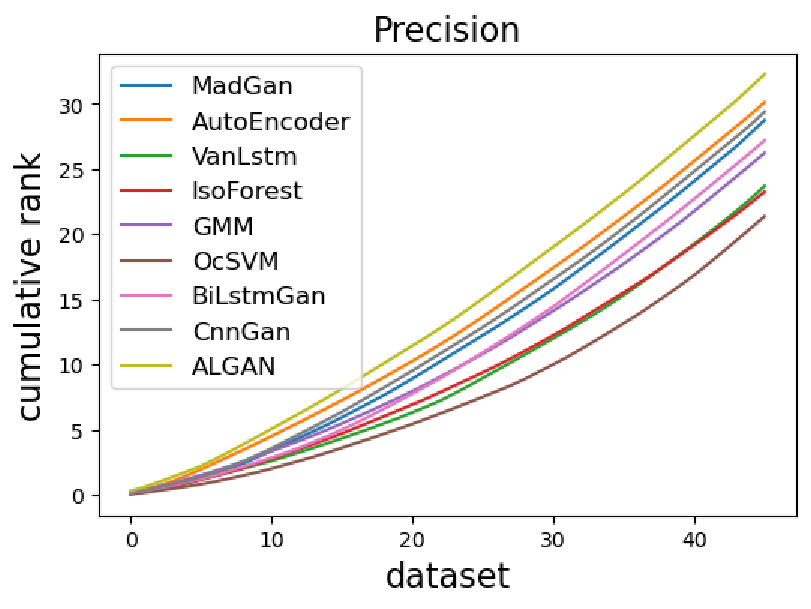}}}
    \quad
    \subfloat{{\includegraphics[width=4cm]{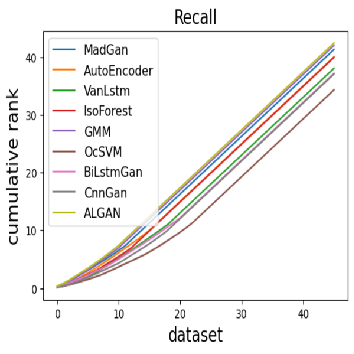}}}
    
    \caption{Cumulative ranking obtained by individually ranking each model per dataset per measure}%
    \label{fig:cr}%
\end{figure*}

Figure \ref{fig:cr} depicts the relative ranking of ALGAN compared to baseline models. The ranking was calculated by aggregating results over all datasets through a cumulative ranking approach based on a comprehensive evaluation criterion. Specifically, to compute the cumulative ranking, we first ranked each method for every dataset based on the evaluation criterion. Ranks were then summed up for each method for a criterion, resulting in a cumulative ranking score. The results demonstrate that ALGAN outperforms all the baseline models for all metrics. 

\begin{figure*}[htb!]
    \centering
    \subfloat{{\includegraphics[width=3cm]{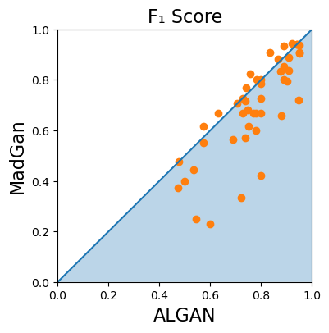}}}
    \quad
    \subfloat{{\includegraphics[width=3cm]{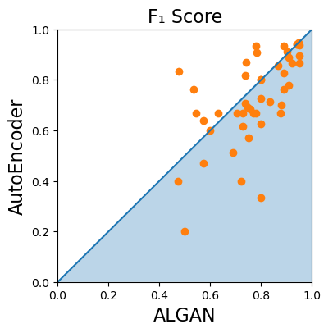}}}
    \quad
    \subfloat{{\includegraphics[width=3cm]{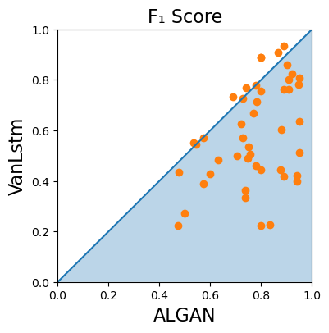}}}
    \quad
    \subfloat{{\includegraphics[width=3cm]{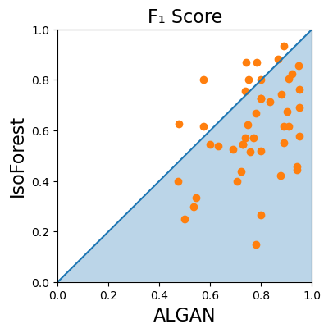}}}
    
    \subfloat{{\includegraphics[width=3cm]{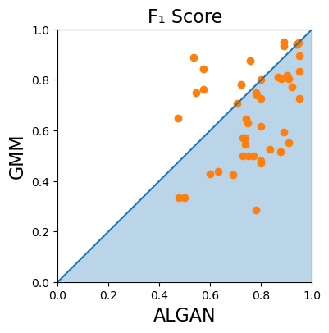}}}
    \quad
    \subfloat{{\includegraphics[width=3cm]{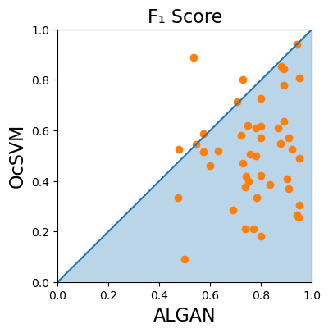}}}
    \quad
    \subfloat{{\includegraphics[width=3cm]{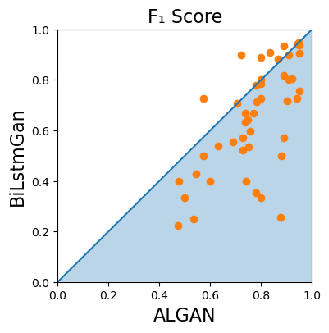}}}
    \quad
    \subfloat{{\includegraphics[width=3cm]{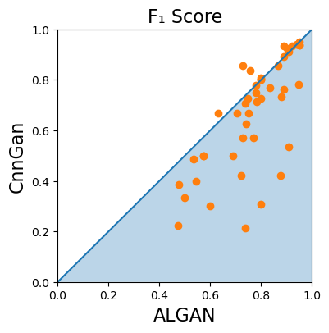}}}

    \caption{Pairwise performance comparison against baseline models. Each dot indicates a dataset. A dot in the blue shade triangle indicates ALGAN performs better than its competitor and dots in the white triangle indicate otherwise, with dots on the diagonal indicating the equal performance of both models.}%
    \label{fig:pwc}%
    \vspace{-.6cm}
\end{figure*}

Figure \ref{fig:pwc} shows the pairwise comparison of ALGAN and each baseline model. The number of dots in the blue shade triangle indicates the number of datasets where ALGAN outperforms a baseline model for the most important measure of anomaly detection namely F$_1$ measure. It can be ascertained that ALGAN outperforms all of the considered baseline models on most of the datasets, highlighting its superior performance in anomaly detection.

\begin{figure}[htb!]
    \centering
    \includegraphics[width=7cm]{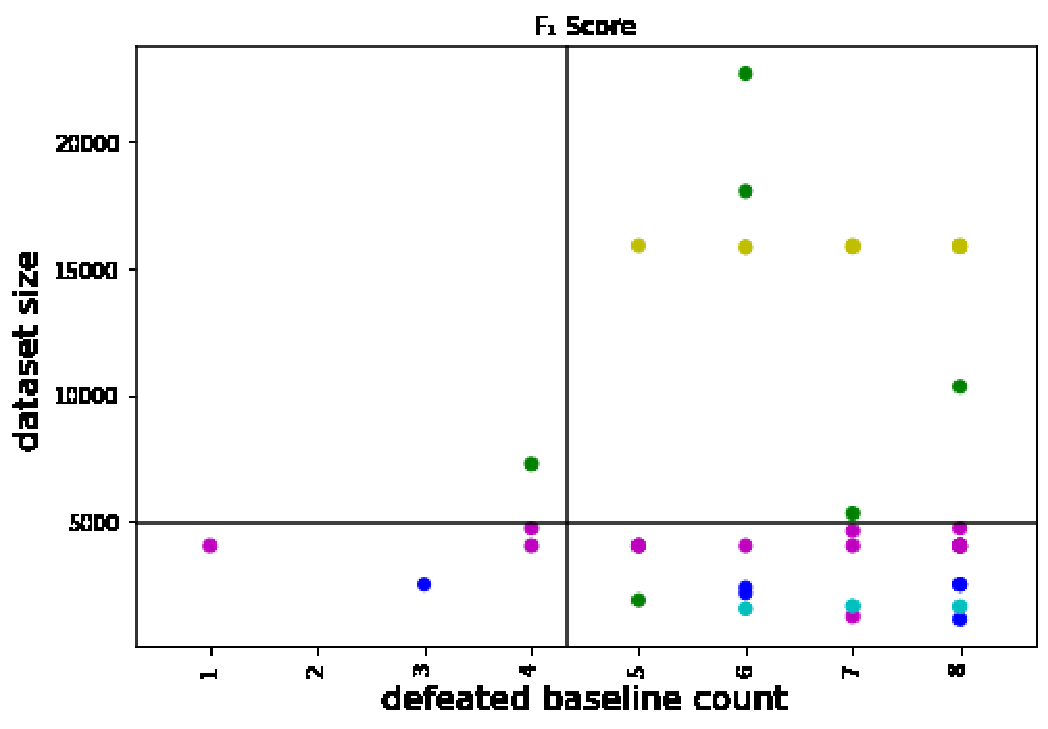}
    \caption{Each point in the scatter plot represents one of the 46 datasets and the colour indicates its membership to one of the five groups (or domain) in the NAB collection. The X-axis shows the number of baseline models outperformed by ALGAN. There are a total of eight baseline models. The Y-axis shows the size of each dataset. }
    \label{fig:beat_dist}
    \vspace{-.6cm}
\end{figure}

Figure \ref{fig:beat_dist} presents an in-depth analysis of model performance to dataset size and domain. The vertical axis represents the size of the dataset and the horizontal axis indicates the number of baseline models that ALGAN outperforms. Each unique colour dot corresponds to a specific category in NAB, indicating the dataset domain. Notably, the figure encompasses all 46 datasets, indicating that ALGAN never performs worse than at least one of the compared algorithms. The majority of the dots are situated towards the right-hand side of the diagram, indicating that ALGAN outperforms the majority of the models. Furthermore, most of the dots appear towards the lower side of the diagram, suggesting that ALGAN excels in small dataset sizes. Specifically, when the dataset size is less than 5000 instances, ALGAN outperforms most models in most datasets, except for a few instances that require further improvement. 
Moreover, when the dataset size is greater than 5000 instances, ALGAN consistently outperforms the majority of the models. Importantly, the distribution of the same colours on both the left and right sides of the diagram demonstrates that ALGAN is not sensitive to the domain of the data collection.

\begin{figure*}[htb!]
    \centering
    
     \subfloat{{\includegraphics[width=8cm]{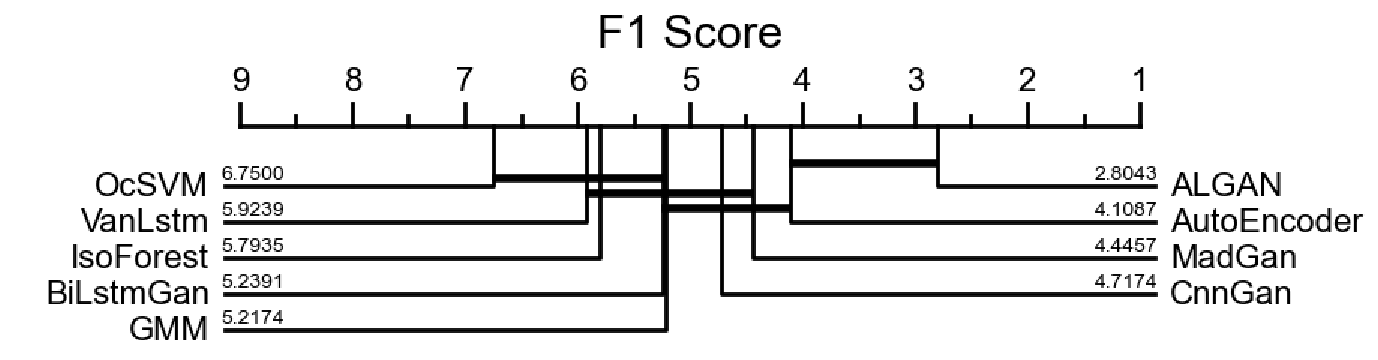}}}
    \quad
    \subfloat{{\includegraphics[width=8cm]{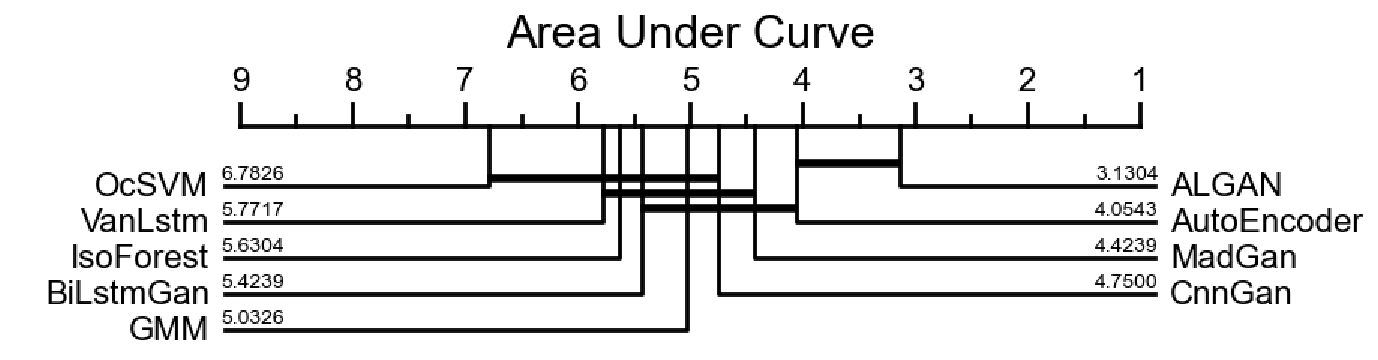}}}
    
    \subfloat{{\includegraphics[width=8cm]{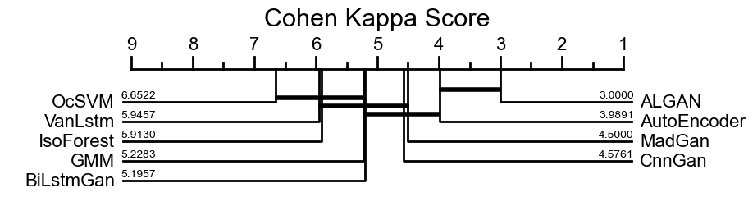}}}
    \quad
    \subfloat{{\includegraphics[width=8cm]{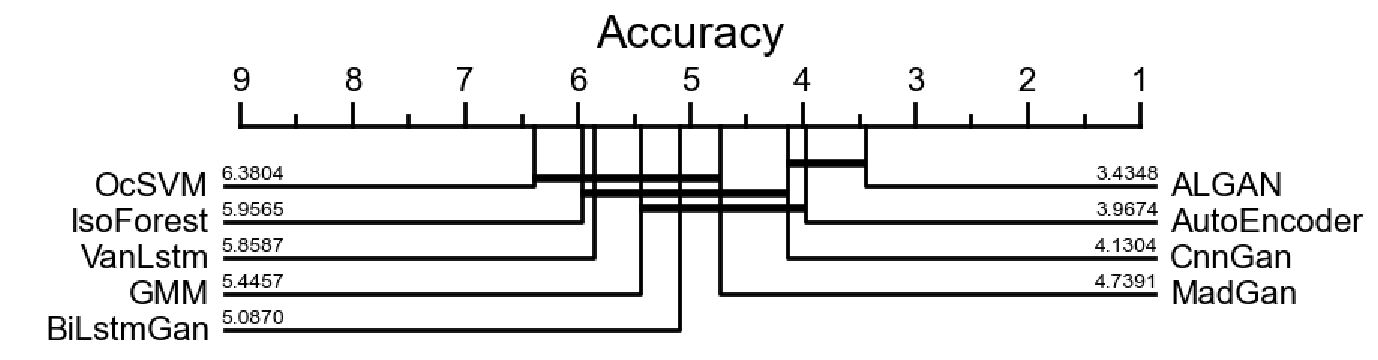}}}
    \quad
    \subfloat{{\includegraphics[width=8cm]{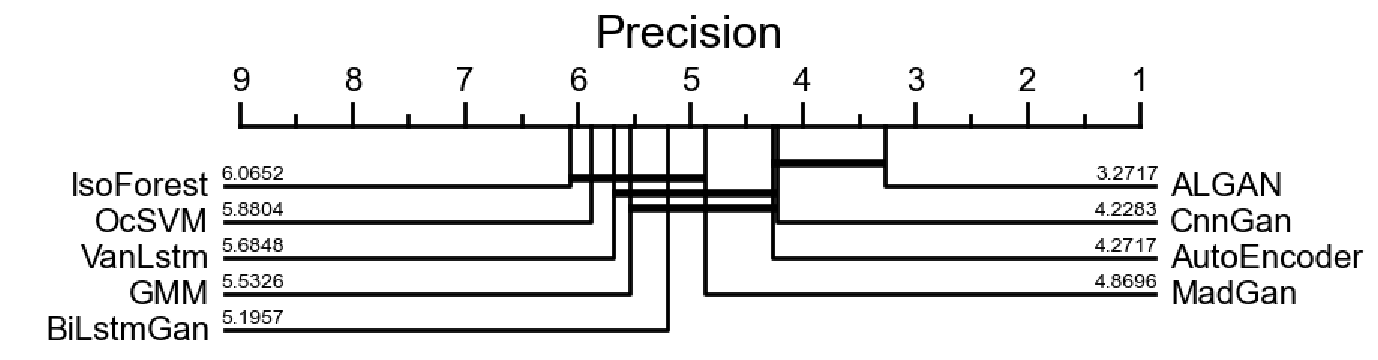}}}
    \quad
    \subfloat{{\includegraphics[width=8cm]{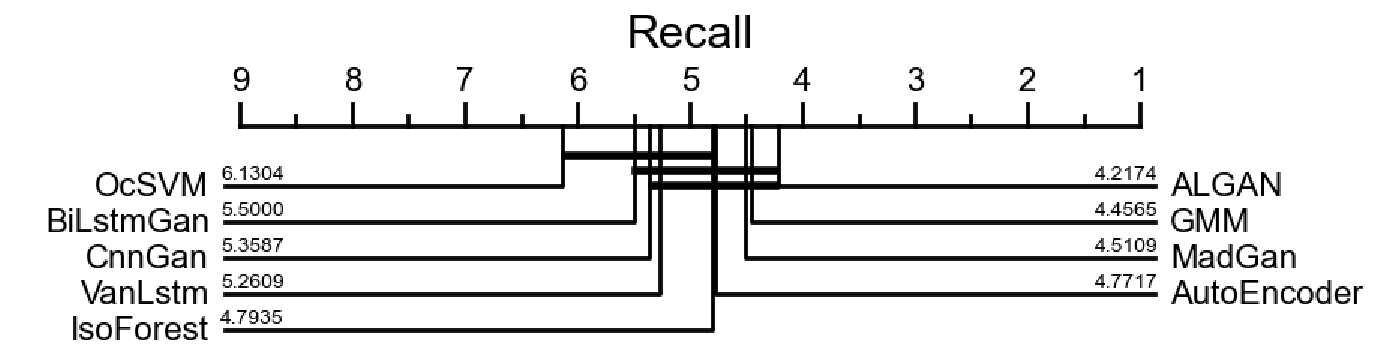}}}
    
    \caption{The critical difference diagram built with all models on various measures}%
    \label{fig:cd}%
    \vspace{-.6cm}
\end{figure*}

Figure \ref{fig:cd} shows the critical difference diagram, which is generated by evaluating each method on a particular evaluation criterion per dataset. The critical difference has been proven to be an effective tool for ranking the performance of models, particularly when the number of datasets for evaluation is high \cite{IsmailFawaz2019DeepReview}. In this diagram, the models are arranged on a horizontal line in ascending order of their scores, where a lower score indicates superior performance \cite{Demsar2006StatisticalSets}.

The critical difference diagram reveals that ALGAN exhibits the lowest score, thereby demonstrating its superior performance over all other models. 
Although ALGAN demonstrates only a marginally better recall than GMM, its poor $F_1$ scores compared with ALGAN imply that GMM produces many false positives, resulting in low precision (as indicated in the precision diagram). Although better recall is desirable in anomaly detection, excessive false positives will lead to numerous false alarm investigations. Thus, a balance between recall and precision is required, as indicated by a higher $F_1$ score, which is achieved by ALGAN.

\begin{table*}[htbp]
  \centering
  \caption{Results on NAB Data collection, results averaged over all 46 datasets in the collection (bold means best result).}
    \begin{tabular}{lcccccc}
    \toprule
          & Accuracy & Precision & Recall & F1    & CK    & AUC \\
    \midrule
    VanLstm & 0.815 & 0.515 & 0.827 & 0.586 & 0.498 & 0.819 \\
    GMM & 0.846 & 0.570 & 0.913 & 0.672 & 0.594 & 0.872 \\
    OcSVM & 0.776 & 0.465 & 0.745 & 0.513 & 0.412 & 0.762 \\
    AutoEncoder & 0.901 & 0.655 & 0.870 & 0.725 & 0.669 & 0.887 \\
    IsoForest & 0.826 & 0.506 & 0.868 & 0.606 & 0.522 & 0.844 \\
    MadGan & 0.857 & 0.591 & 0.808 & 0.649 & 0.576 & 0.834 \\
    BiLstmGan & 0.859 & 0.625 & 0.897 & 0.696 & 0.630 & 0.875 \\
    CnnGan & 0.882 & 0.639 & 0.805 & 0.674 & 0.614 & 0.850 \\

    ALGAN & \textbf{0.915} & \textbf{0.701} & \textbf{0.922} & \textbf{0.771} & \textbf{0.724} & \textbf{0.917} \\
    \bottomrule
    \end{tabular}%
  \label{tab:addlabel}%
\end{table*}%

\subsection{Accuracy Performance on SWaT Dataset}
The experimental results in Table \ref{tab:SWaT_results} demonstrate that ALGAN outperforms all other baseline models, suitable for multivariate datasets, in terms of recall, F1 score, CK and AUC metrics. Although AutoEncoder achieves the same accuracy as ALGAN and slightly higher precision than ALGAN, it suffers from lower recall implying that it fails to detect many true positives. Comparatively, ALGAN provides a more balanced performance as shown by yielding the highest F1 score. A balanced trade-off between precision and recall, reflected by F1 score, is desirable for anomaly detection tasks.
\begin{table*}[htbp]
  \centering
  \caption{Performance on SWaT dataset (bold means best result) with models that can be run on multivariate datasets}
    \begin{tabular}{lcccccc}
    \toprule
          & Accuracy & Precision & Recall & F$_1$    & CK    & AUC \\
    \midrule
    AutoEncoder & \textbf{0.966} & \textbf{0.968} & 0.752 & 0.846 & 0.827 & 0.874 \\
    BiLstmGan & 0.962 & 0.945 & 0.744 & 0.833 & 0.812 & 0.869 \\
    MadGan & 0.962 & 0.927 & 0.759 & 0.835 & 0.814 & 0.875 \\
    CnnGan & 0.961 & 0.947 & 0.729 & 0.824 & 0.802 & 0.862 \\
    ALGAN & \textbf{0.966} & 0.952 & \textbf{0.769} & \textbf{0.851} & \textbf{0.832} & \textbf{0.882} \\
    \bottomrule
    \end{tabular}%
  \label{tab:SWaT_results}%
\end{table*}%

\subsection{Discussion}
The aforementioned results provide empirical evidence to support the efficacy of GANs in detecting anomalies within time series data of both univariate and multivariate nature. In particular, the generative model's ability to learn the general distribution of the dataset has proven useful in isolating anomalous instances from normal ones. It also shows the ability of ALstm to reduce the forgetting problem when handling long time series or sequences (Figure \ref{fig:beat_dist}). 

MadGan \cite{Li2019}, BiLstmGan \cite{Zhu2019ElectrocardiogramNetwork}, and CnnGan \cite{Schlegl2017UnsupervisedDiscovery} have a similar architecture, i.e. GAN, as ALGAN. However, ALGAN outperforms these models on both NAB and SWaT datasets, as shown by various metrics. Distinct from these state-of-the-art models, ALGAN uses ALstm as an underlying model in both the generator and discriminator. 
The attention layer on top of the input and the LSTM output allows ALstm to decode the information more accurately by reducing the information loss in the propagation of processing time series and sequence data. This architecture enables ALstm to focus on important time stamps and update those steps to reduce information loss. This is especially beneficial for small datasets (e.g. NAB collection) that do not have enough instances, where information loss can significantly affect the results. By reducing information loss, ALGAN achieves significantly better results than the baseline models on NAB as well as on the large dataset SWaT.

Among the anomaly detection models, AutoEncoder is the second most effective method for both NAB and SWaT data collections. AutoEn employs an autoencoder architecture that compresses the time series data into a lower-dimensional latent space and then reconstructs the data into a higher-dimensional output space. By doing so, AutoEncoder can capture the general distribution of the data better than other models such as LSTM or CNN. For example, VanLstm, a vanilla LSTM model only uses the hidden states of the LSTM cells to detect anomalies. VanLstm suffers from the limitations of LSTM networks, such as gradient vanishing and exploding problems \cite{Malhotra2015,hochreiter1997long}. In contrast, AutoEncoder leverages the compressed representation from bottleneck layer to learn more meaningful and compact features from the input data \cite{Borghesi2019AnomalySystems,Zhou2017AnomalyAutoencoders}. 

However, AutoEncoder has some drawbacks such as they may be lossy, which means that they may lose some information or quality in the compression and reconstruction process. During this process, they try to to capture as much information as possible rather than as much relevant information as possible. This means that they may not learn the most useful features for a specific task or goal.  This may affect the performance or accuracy of the model for some tasks that require high fidelity or precision. These issues are addressed by ALGAN using a novel anomaly detection model that combines ALstm and adversarial training. ALstm, an improved version of LSTM, incorporates attention mechanisms to enhance the temporal relationships among the hidden states and the inputs and to capture the most relevant features. Adversarial training enables the generator to learn a more faithful representation of the data distribution instead of only learning specific parts. Results in previous sections demonstrate the superiority of ALstm and adversarial training for anomaly detection in time series data. 

Among the anomaly detection models, OcSVM \cite{Scholkopf} is the least effective method. It is based on a very popular one-class support vector machine that tries to find a hyperplane that separates the normal data points from the origin. Since it relies on a linear kernel, it fails to capture the complex and nonlinear patterns present in the data. Therefore, OcSVM yields a low accuracy and a high false positive rate. 

Belonging to the decision tree family, IsoF \cite{Liu2008IsolationForest} isolates instances based on their features. It randomly selects a feature and a split value for each node of a tree and recursively partitions the data until each instance is isolated. It assumes that anomalies are more likely to be isolated earlier than normal instances. However, it has a limitation that it cannot capture the nonlinear interaction between the features, which is crucial for time series data. 

GMM \cite{bishop2006linear,Reynolds2009} fits a mixture of Gaussian distributions to the data to learn the covariance structure of the data distribution, which can be arbitrary and complex. By doing so, GMM can capture the variations and patterns in the data more accurately than other traditional (non-neural network) models. Therefore, GMM achieves significantly better results than those models on various metrics. However, GMM is still inferior to ALGAN.

In summary, empirical analysis ascertains that ALGAN can learn complex data distribution and non-linear relationships present in the data. ALGAN does it by leveraging the attention mechanisms of ALstm to enhance the temporal dependencies among the inputs and the hidden states, and the competitive training of GANs to learn a more faithful representation of the data distribution. We evaluated ALGAN on two collections: NAB and SWaT, which are widely used for anomaly detection in time series data. We compared ALGAN with several popular unsupervised anomaly detection models, including traditional models such as OcSVM, IsoF, and GMM, and neural network-based models such as VanLstm, AutoEn, MadGan, BiLstmGan, and CnnGan. The results showed that ALGAN consistently outperformed all the baseline models on various metrics, demonstrating the superiority of ALstm and adversarial training for anomaly detection in time series data. Overall, ALGAN represents a promising approach for detecting anomalies in time series data, with potential applications in various domains such as finance, healthcare, and cybersecurity.

\section{Conclusion}
\label{sec:conclusions}
This paper presents a novel technique ALGAN for detecting anomalies in both univariate and multivariate time series data. ALGAN leverages an Adjusted Long Short-Term Memory in a Generative Adversarial Network architecture. The approach involves training an Adjusted-LSTM-based generator to learn the general data distribution of the data and then projecting time sequences into a latent space using inverse mapping. Anomaly scores are estimated by using the reconstruction loss of the sequences as they are reconstructed by the generator. By incorporating Adjusted-LSTM into both the generator and discriminator of the GAN, ALGAN improves the accuracy of anomaly detection for both types of datasets.

The experimental results on diverse real-world time series datasets, including 46 univariate and one large multivariate datasets, demonstrate that ALGAN outperforms traditional models, neural network-based models, and other GAN-based models for anomaly detection in time series. However, there are two known challenges for GAN-based anomaly detection in time series: determining an optimal window length and addressing model instability. Additionally, the sensitivity of GAN-based anomaly detection to the number of epochs (i.e. training iteration) is a factor that warrants further investigation. In future, we plan to assess ALGAN on other large multivariate datasets and compare it against additional baseline models to better understand these issues and enhance the performance of ALGAN.

\section*{Declarations}
\noindent \textbf{Funding:}
This work was supported by Verton, a load-management system developing company, under Innovative Manufacturing CRC grant.

\bigskip

\noindent \textbf{Conflicts of Interests:}
We have no other conflicts of interest or competing interests to declare.

\bigskip

\noindent \textbf{Human Participants and/or Animals:}
This research did not involve any human participants and/or animals.

\bigskip

\noindent \textbf{Informed Consent:}
Authors and associated organisations were fully informed and gave consent about this research and publication.

\bigskip

\noindent \textbf{Author Contributions:}
The project was designed and supervised by Richi Nayak. The technical solution was devised by Md Abul Bashar and Richi Nayak. Bashar developed models and conducted experiments. Both authors wrote and reviewed the manuscript.


\bibliographystyle{compj}
\bibliography{references}

\end{document}